\documentclass[10pt,journal,final]{IEEEtran}

\usepackage{cite}
\usepackage{url}
\usepackage[dvipsnames]{xcolor}
\usepackage{bbding}
\usepackage{hyperref}

\usepackage[pdftex]{graphicx}
\graphicspath{{./figures/}} 
\usepackage[caption=false,font=footnotesize]{subfig}
\usepackage{overpic}
\usepackage{ragged2e}

\usepackage{amsmath}
\usepackage{array}
\usepackage{multirow}
\usepackage[ruled,linesnumbered]{algorithm2e}
\usepackage{mathrsfs}
\usepackage{amssymb}
\usepackage{makecell}

\usepackage{pifont}
\newcommand{\cmark}{\ding{51}}%
\newcommand{\xmark}{\ding{55}}%

\def\eg{\emph{e.g.}} 
\def\ie{\emph{i.e.}} 
\def\etal{\textit{et~al.}} 

\begin{document}

\title{TMP: Temporal Motion Propagation for\\ Online Video Super-Resolution}

\author{Zhengqiang~Zhang,
        Ruihuang~Li,
        Shi~Guo,
        Yang~Cao,
        and Lei~Zhang
\thanks{Z.~Zhang, R.~Li, S.~Guo, and L.~Zhang are with the Department of Computing, The Hong Kong Polytechnic University, Hong Kong (e-mail: zhengqiang.zhang@connect.polyu.hk; csrhli@comp.polyu.edu.hk; guoshi28@outlook.com; cslzhang@comp.polyu.edu.hk). This work is supported by the Hong Kong RGC RIF grant (R5001-18) and the PolyU-OPPO Joint Innovation Lab.}
\thanks{Y.~Cao is with the Department of Computer Science and Engineering, The Hong Kong University of Science and Technology (e-mail: yangcao.cs@gmail.com).}}



\maketitle


\begin{abstract}
Online video super-resolution (online-VSR) highly relies on an effective alignment module to aggregate temporal information, while the strict latency requirement makes accurate and efficient alignment very challenging.
Though much progress has been achieved, most of the existing online-VSR methods estimate the motion fields of each frame separately to perform alignment, which is computationally redundant and ignores the fact that the motion fields of adjacent frames are correlated.
In this work, we propose an efficient Temporal Motion Propagation (TMP) method, which leverages the continuity of motion field to achieve fast pixel-level alignment among consecutive frames. 
Specifically, we first propagate the offsets from previous frames to the current frame, and then refine them in the neighborhood, which significantly reduces the matching space and speeds up the offset estimation process. 
Furthermore, to enhance the robustness of alignment, we perform spatial-wise weighting on the warped features, where the positions with more precise offsets are assigned higher importance.  
Experiments on benchmark datasets demonstrate that the proposed TMP method achieves leading online-VSR accuracy as well as inference speed.
The source code of TMP can be found at \href{https://github.com/xtudbxk/TMP}{https://github.com/xtudbxk/TMP}.
\end{abstract}


%
\IEEEpeerreviewmaketitle


\section{Introduction}
\IEEEPARstart{V}{ideo} super-resolution (VSR) aims to reconstruct a high-resolution (HR) video sequence from its low-resolution (LR) counterpart. In recent years, due to the increasing popularity of online video conference, live broadcasting, \textit{etc.}, online-VSR has attracted growing attention from researchers~\cite{dap,ckbg,r2d2}. In online-VSR, the enhancement of the current frame is achieved by using only the information from past frames in an online manner.  Compared with other VSR problems, the stricter constraint on latency of online-VSR brings more challenges to frame alignment, which is critical for the efficiency of a VSR model.

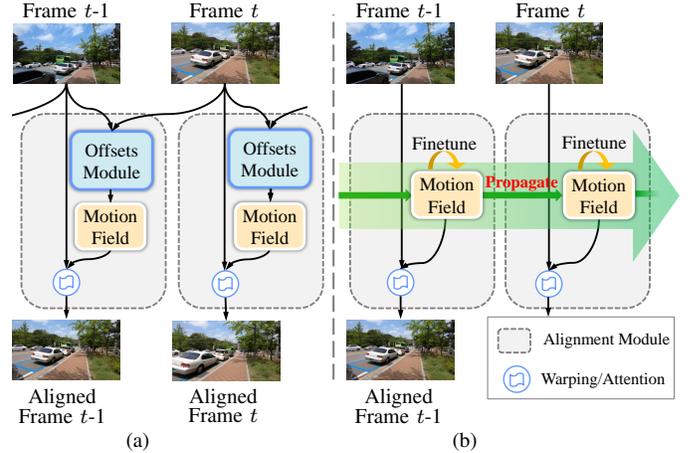
\begin{figure}[!t]
\centering
\begin{overpic}[width=\linewidth]{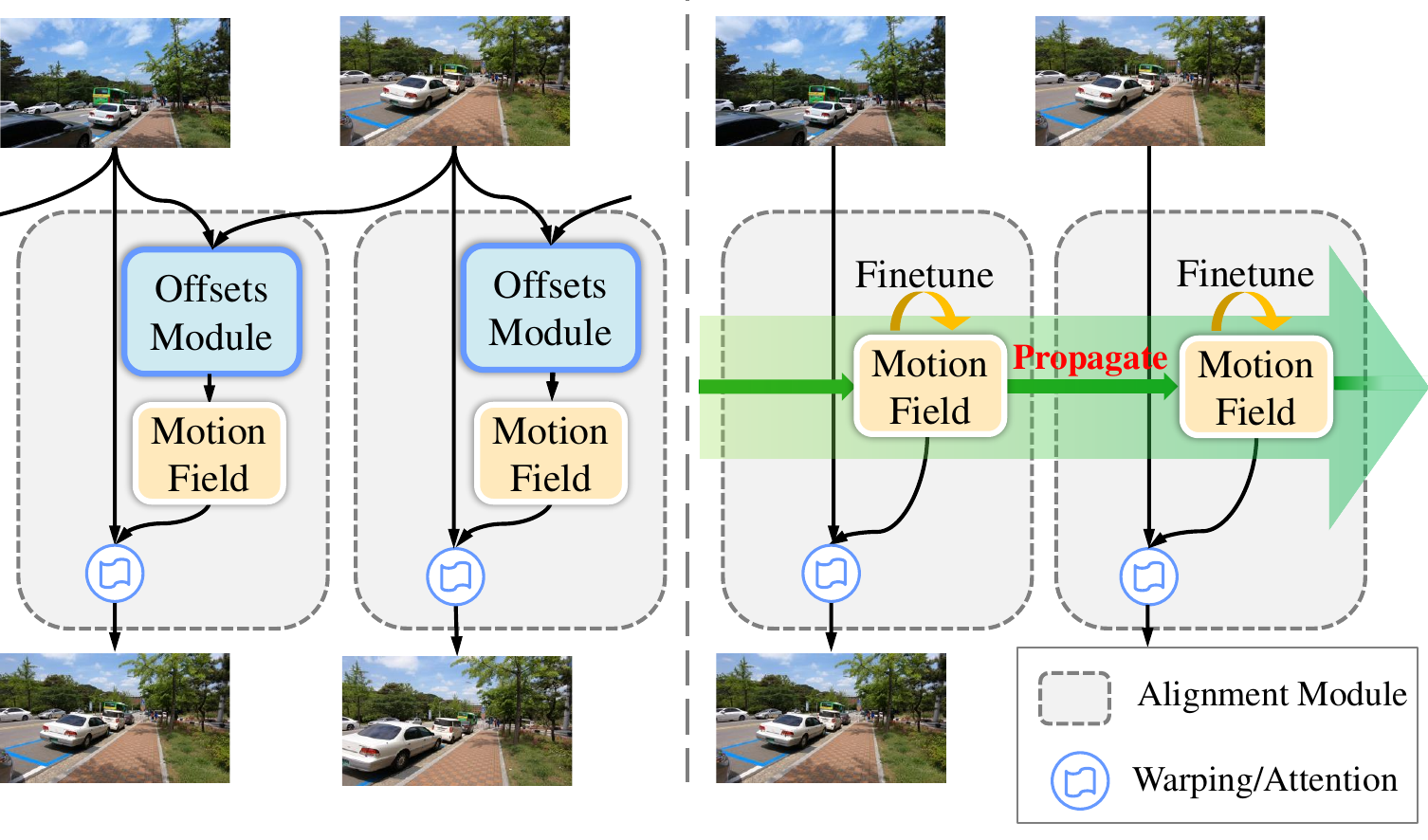}
\put(1.5,58.5){\footnotesize{Frame $t$-1}}
\put(26.5,58.5){\footnotesize{Frame $t$}}
\put(51.5,58.5){\footnotesize{Frame $t$-1}}
\put(75.5,58.5){\footnotesize{Frame $t$}}

\put(-1.5,-1){\fontsize{8}{6}\selectfont {\begin{tabular}{c}Aligned \\ Frame $t$-1\end{tabular}}}
\put(24,-1){\fontsize{8}{6}\selectfont {\begin{tabular}{c}Aligned \\ Frame $t$\end{tabular}}}
\put(49,-1){\fontsize{8}{6}\selectfont {\begin{tabular}{c}Aligned \\ Frame $t$-1\end{tabular}}}

\put(17,-6){\footnotesize{(a)}}
\put(66,-6){\footnotesize{(b)}}
\end{overpic}
\vspace{0.5pt}

\caption{(a) Existing online-VSR methods compute the motion field of each frame separately via a weight-sharing offsets estimation module, while (b) our method propagates the motion field of previous frame to the current frame, reducing much the computational cost.}
\label{fig:motivation}
\end{figure}

Compared to single image super-resolution (SISR) ~\cite{srcnn, fsrcnn, edsr, esrgan, dai2019second, MSPB, DnCNN }, VSR employs an additional alignment module to aggregate the complementary information from adjacent frames for achieving higher reconstruction ability and better visual quality on the current frame.
As a fundamental part of VSR models, a few frame alignment methods~\cite{detail_revealing, robustvsr, frvsr, tdan, edvr, pfnl, mucan, duf, STAN, LTDVSR} have been developed to improve the performance of VSR methods.
Some methods~\cite{tdan, understanding, edvr, basicvsr++} adopt the deformable strategy to predict the offsets for every pixel, while some methods ~\cite{pfnl, mucan, mana} enlarge the searching range by using a non-local attention module to handle large motions.
Though having achieved considerable improvement in VSR performance, these methods can hardly be applied to online-VSR tasks due to their long latency and high complexity. 
To solve this issue, DAP~\cite{dap} employs a deformable attention pyramid module, which dynamically predicts a limited number of offsets for each pixel and merges the corresponding features in a weighted average manner.
On the other hand, FRVSR~\cite{frvsr} and CKBG~\cite{ckbg} resort to the optical flow for frame alignment.
They adopt a lightweight optical flow network to estimate the displacement between two frames and align them by warping.
%
%
Nonetheless, those methods still waste much computation on the repeated estimation of offsets.
As shown in Figure~\ref{fig:motivation}(a), most of the existing VSR methods apply a shared alignment module to each frame and compute their motion fields separately, ignoring the fact that the motion fields of neigboring frames are highly correlated.
In this work, we propose an efficient \textbf{Temporal Motion Propagation} (\textbf{TMP}) method to speed up the alignment process by replacing the complex offsets estimation module with an efficient motion field temporal propagation scheme.
As illustrated in Figure~\ref{fig:motivation}(b), instead of computing the motion field of each frame from scratch, we inherit the motion field from the previous frame, and apply a fast finetuning process to refine it.
Considering the different behaviors of different types of motions, we design two propagation paths. One is the object motion propagation path, which locates the potential destinations of \textbf{moving} objects in the current frame, while the other is the camera motion propagation path, which exploits the position information of \textbf{static} regions in the previous frame. The two paths work together to produce a few offset candidates for each pixel in the current frame. 
In the finetuning step, we pass the offsets of each pixel to its neighbors and evaluate all the candidates to pick the best one that has the least matching distance. Finally, we use the refined motion field to perform frame alignment by a warp operation.

The alignment process will inevitably introduce some noise (\ie, errors) into the motion fields, which should be alleviated in the feature fusion process.
Some approaches~\cite{vsrnet,edvr, iam} employ the reweighting strategy to assign different importance to the aligned features in a spatial-wise manner according to their similarities to the reference features.
However, in this way some well-matched features which contain complementary textures will be discarded since they may have a low similarity to the reference features.
To address this issue, we propose to weigh the aligned features based on the confidence of estimated offsets instead of the similarities between texture features.
Specifically, we employ two separate branches to extract motion and texture features, respectively, and use only the motion features to evaluate the precision of motion fields. 
In this way, we can more effectively assess the confidence of motion fields since it is untangled from textures.

Extensive ablation studies and experiments on benchmark datasets demonstrate the effectiveness and efficiency of our TMP method for online-VSR.
Compared with the existing state-of-the-arts, our model achieves not only higher accuracy but also faster inference speed. In particular, compared with DAP \cite{dap}, our model achieves a PSNR gain of $0.08$ dB with nearly 65\% latency on the REDS4 dataset~\cite{reds}.

The rest of this paper is organized as follows.
Section~\ref{sec:2} summarizes the relevant works on offline/online-VSR and feature alignment.
Section~\ref{sec:3} presents the proposed TMP algorithm in detail.
Section~\ref{sec:4} reports the experimental results and analysis.
Finally, Section~\ref{sec:5} concludes this paper.

\begin{figure*}[!t]
\centering
\begin{overpic}[width=\linewidth]{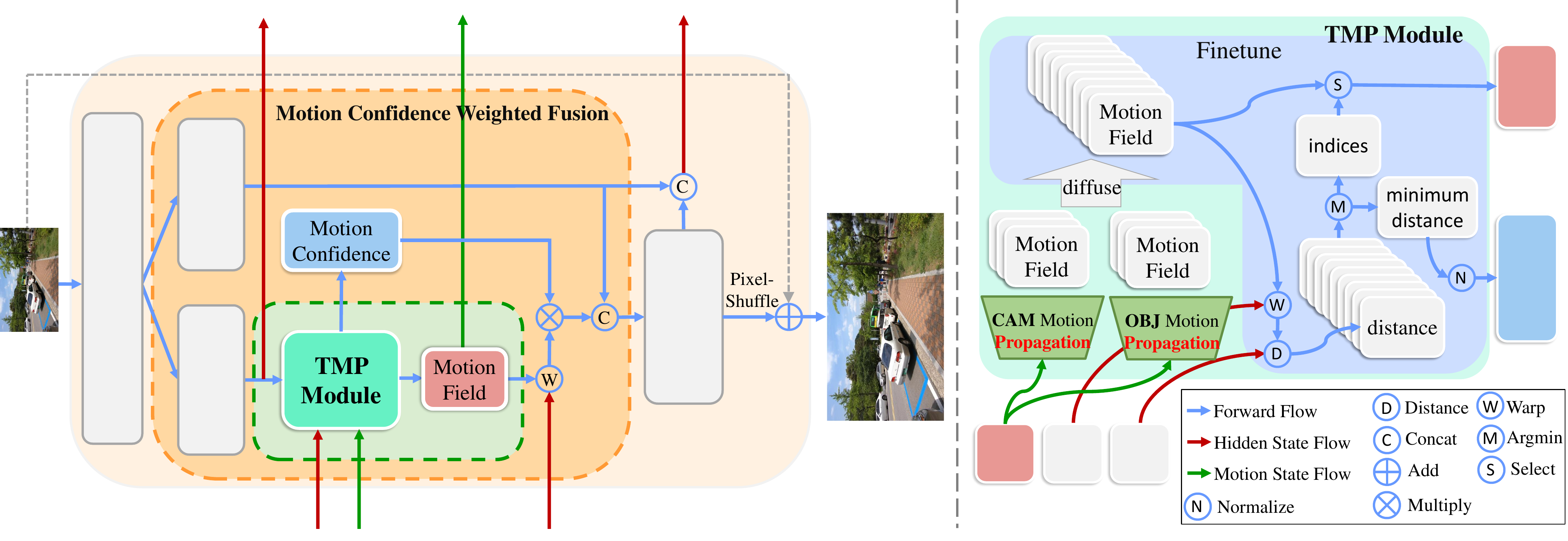}
\put(6.5, 11.3){\rotatebox{90}{\fontsize{7}{8}\selectfont Feature Extraction}}
\put(13, 5.9){\rotatebox{90}{\fontsize{7}{8}\selectfont Motion Branch}}
\put(13, 17.6){\rotatebox{90}{\fontsize{7}{8}\selectfont Texture Branch}}
\put(43, 10){\rotatebox{90}{\fontsize{7}{8}\selectfont Reconstruction}}
\put(49, 18.5){\rotatebox{90}{\fontsize{6}{8}\selectfont Bilinear Upsampling}}
\put(95.8, 12.3){\rotatebox{90}{\fontsize{7}{8}\selectfont \begin{tabular}{c} Motion \\ Confidence\end{tabular}}}
\put(95.8, 25.6){\rotatebox{90}{\fontsize{7}{8}\selectfont \begin{tabular}{c}Motion \\ Field\end{tabular}}}

\put(16.5, 1.5){\fontsize{7}{8}\selectfont $H^0_{t-1}$}
\put(23, 1.5){\fontsize{7}{8}\selectfont $M_{t-1}$}
\put(35.2, 1.5){\fontsize{7}{8}\selectfont $H^1_{t-1}$}
\put(14.2, 31.5){\fontsize{7}{8}\selectfont $H^0_t$}
\put(27, 31.5){\fontsize{7}{8}\selectfont $M_t$}
\put(41, 31.5){\fontsize{7}{8}\selectfont $H^1_t$}

\put(62.2, 5.3){\fontsize{6}{8}\selectfont $M_{t-1}$}
\put(66.55, 5.3){\fontsize{6}{8}\selectfont $H^0_{t-1}$}
\put(71.6, 5.3){\fontsize{6}{8}\selectfont $H^0_t$}

\end{overpic}

\caption{Overview of our proposed online-VSR method. Left: The flowchart of the proposed method. There are two major differences between our method and the existing methods. One is the temporal motion propagation (TMP) module (highlighted in \textcolor{BlueGreen}{green} color box), which propagates the motion field from the previous frame to the current frame. The other is the motion confidence weighted fusion (highlighted in \textcolor{Orange}{orange} color box), which weighs the warped features by the accuracy of estimated offsets. Right: The detailed architecture of the TMP module. Best viewed in color.}
\label{fig:framework}
\end{figure*}

\section{Related Work} \label{sec:2}

\subsection{Video Super-resolution}
Video super-resolution (VSR) aims at reconstructing visually pleasing HR videos from the LR inputs via exploiting the temporal dependency between consecutive frames.
Depending on the usage of supporting frames, existing VSR methods can be divided into four groups.
Most of the earlier works~\cite{tdan, edvr,pfnl, mucan, tga, toflow, duf, MMCNN, SOFSR, MSFN, STAN, DDAN} belong to the \textit{sliding-window based methods}, which employ the frames within a short temporal window to enhance the center frame. 
Xue~\etal~\cite{toflow} warped the neighboring frames to the reference frame by using task-oriented motion cues, acheiving obvious improvement.
Song~\etal~\cite{MSFN} proposed to aggregate the sequential information at different stages of the neural network. 
%
%
TDAN~\cite{tdan} introduces deformable convolutions to align the adjacent frames at the feature level, which is further extended by EDVR~\cite{edvr} using a pyramidal cascade structure to fuse the temporal information at various scales.
%
%
The second group is the \textit{bidirectional propagation methods} \cite{basicvsr, basicvsr++, RRCN}, which leverage the information of past, current and future frames in the sequence to enhance each frame.
For example, RRCN~\cite{RRCN} utilizes two single-direction recurrent convolutional networks to collect the complementary information from the whole video, while BasicVSR~\cite{basicvsr} and BasicVSR++~\cite{basicvsr++} adopt a typical bidirectional loop structure as backbone.
Due to a large amount of supporting frames, these methods typically achieve better accuracy by compromising the latency.
The third group is unidirectional propagation methods \cite{rlsp, etdm}, which aggregate the information from the past and current frames, as well as several cached future frames.
For example, RSLP~\cite{rlsp} transmits the information from the supporting frames using high dimensional latent states.
ETDM~\cite{etdm} utilizes the buffer mechanisms to store differences between consecutive frames.
When only the past and current frames can be used, unidirectional propagation methods become \textit{online-VSR methods}~\cite{frvsr,rsdn,rrn, dap}, which have recently attracted much research attention due to their short latency and hence potential applications in online video conferences and live broadcasting. 

\subsection{Online Video Super-resolution}
The key issue of online-VSR is its strict latency requirement.
One way to handle this problem is to directly apply SISR methods~\cite{Cubic, RFI, SI, USVSR, fastSRCNN, classsr, etdm}.
For example, the interpolation-based SR approaches are widely applied in the online video processing tasks, providing marginal accuracy improvement but high speed.  
Zhang~\etal~\cite{RFI} developed a bivariate rational fractal interpolation model to preserve the textual and structural information of images.
Choi~\etal~\cite{SI} unified the interpolation step and quality-enhancement step to reduce complexity.
In addition, some lightweight learning-based methods have been introduced to solve the online-VSR problem.
Dong~\etal~\cite{fastSRCNN} decreased the resolution of feature extraction modules, greatly reducing the inference latency.
Kong~\etal~\cite{classsr} first classified the image regions to groups of different restoration difficulties, then processed those patches by networks of different capacities to save computations.

%

Due to the better visual quality, designing a lightweight online-VSR network draws increasingly attention.
DAP~\cite{dap} employs a deformable attention pyramid module to dynamically attend to limited salient locations of supporting frames, which significantly reduces the computational burden associated with the classical attention mechanism.
FRVSR~\cite{frvsr} and CKBG~\cite{ckbg} adopt a lightweight network to estimate the optical flow and perform motion compensation.
To break the bottleneck brought by the small network capacity, FRVSR employs the recurrent architecture that passes the previously estimated HR frame to the following reconstruction module, while CKBG utilizes a kernel knowledge transfer method to obtain additional information from the teacher network.
Despite the many efforts, most existing online-VSR methods still suffer from low speed in inference, which hinders the widespread of these methods in many applications.

\subsection{Frame Alignment}
Frame alignment is critical to the performance of VSR for its role in aggregating temporal information.
Early works~\cite{toflow, frvsr, detail_revealing} mostly adopt the optical flow technique to perform alignment.
As pointed out in TOF~\cite{toflow}, however, the traditional optical flow methods are not the best choice for estimating the motion fields in VSR. Later VSR methods focus more on implicit alignment.
TDAN~\cite{tdan} performs motion compensation via deformable convolutional layers~\cite{dcn}, which inspires a lot of following works such as EDVR~\cite{edvr} and BasciVSR++~\cite{basicvsr++}.
EDVR employs a pyramid, cascading and deformable alignment module to handle complex motions in a coarse-to-fine manner, while BasicVSR++ leverages optical flow to improve the training stability of deformable alignment.
PFNL~\cite{pfnl} utilizes the classical non-local mechanism to exploit the sequential information from support video frames. MuCAN~\cite{mucan} and MANA~\cite{mana} adopt a similar strategy to avoid explicit motion estimation.
Besides, DUF~\cite{duf} prefers to predict the upsampling filters to reconstruct the HR videos directly.

Unfortunately, most of the existing VSR methods consider only the correlations between two frames when estimating their relative displacement, ignoring the fact that the motion fields of neighboring frames are similar. 
As a result, DAP~\cite{dap} has to limit the search space to a small range when performing alignment. FRVSR~\cite{frvsr} and CKBG~\cite{ckbg} resort to unsound lightweight optical flow networks for estimating the motion fields.
In this paper, we propose to temporally propagate the estimated motion fields across frames, which significantly reduces the latency by removing the repeated computations of estimating offsets for each frame.

\section{Methodology}  \label{sec:3}

\subsection{Overview}

Online-VSR aims to reconstruct the HR reference frame $I^{HR}_t \in \mathbb{R}^{sH\times sW\times C}$ from its corresponding LR frame $I^{LR}_t \in \mathbb{R}^{H\times W\times C}$ and the $N-1$ support frames $I^{LR}_{[t-N:t-1]}$, where $s$ denotes the upsampling factor, $H$ and $W$ denote the height and width of LR frames, $C$ refers to the channel number of each frame, and $t$ represents the timestamp of online video.
Following previous works \cite{frvsr, rsdn, rrn}, we employ the recurrent architecture for its simplicity and efficiency. 
The overall framework of our proposed method is depicted in the left part of Figure~\ref{fig:framework}.
Our network takes the current LR frame $I^{LR}_t$, the estimated motion field of the last frame $M_{t-1}$ and the hidden states $H^0_{t-1}$, $H^1_{t-1}$ as inputs, and outputs the corresponding HR frame $I^{HR}_t$ and the updated hidden states $H^0_t$ and $H^1_t$.
The hidden state $H^0_t$ is used to refine the inherited motion field, while the hidden state $H^1_t$ stores the texture information from previous video frames.
Compared with existing online-VSR methods, which separately estimate the motion field of each frame for alignment, our proposed network architecture has two major differences.
First, our network forward propagates the estimated motion field across frames to reduce repetitive computation by introducing the Temporal Motion Propagation (TMP) module, whose detailed structures are described in Section~\ref{tmp}.
Second, we use two different branches to process the motion features and texture features, respectively, bringing additional advantages in fusing the aligned features. The details are presented in Section~\ref{mcwf}.

\definecolor{myblue}{RGB}{0, 176, 240}
\definecolor{myorange}{RGB}{255, 164, 5}
\definecolor{myred}{RGB}{204, 5, 5}
\definecolor{mygreen}{RGB}{63, 195, 125}

\subsection{Temporal Motion Propagation} \label{tmp}

Considering the temporal smoothness of motions, we propose a novel TMP module to inherit the estimated motion field from the last frame and finetune it based on sequential contexts.
TMP circumvents the problem of motion field estimation from scratch for each frame, and hence reduces greatly the inference latency.

Given the hidden states of current frame $H^{0}_{t}$ and past frame $H^{0}_{t-1}$ as inputs, TMP estimates the current motion field $\boldsymbol{M}_t \in \mathbb{R}^{2\times H\times W}$ by taking $\boldsymbol{M}_{t-1}$ of previous frame as a priori. The detailed process of TMP is illustrated in the right part of Figure~\ref{fig:framework}.
We leverage the continuity of motions to propagate the motion information from previous frames to the current frame, and predict a set of offset candidates for $M_t$ dynamically based on the propagated motions. To handle distinct motion behaviors of moving objects and static regions, we employ two different motion propagation paths, \eg, object (OBJ) motion path and camera (CAM) motion path.
During the subsequent finetuning step, we shuffle the candidates of each pixel to its neighbors and find the optimal offset with the least matching distance. 

\begin{figure}
\centering

\subfloat[OBJ motion propagation]{\hspace{-0.4cm}
\begin{overpic}[width=1.06\linewidth]{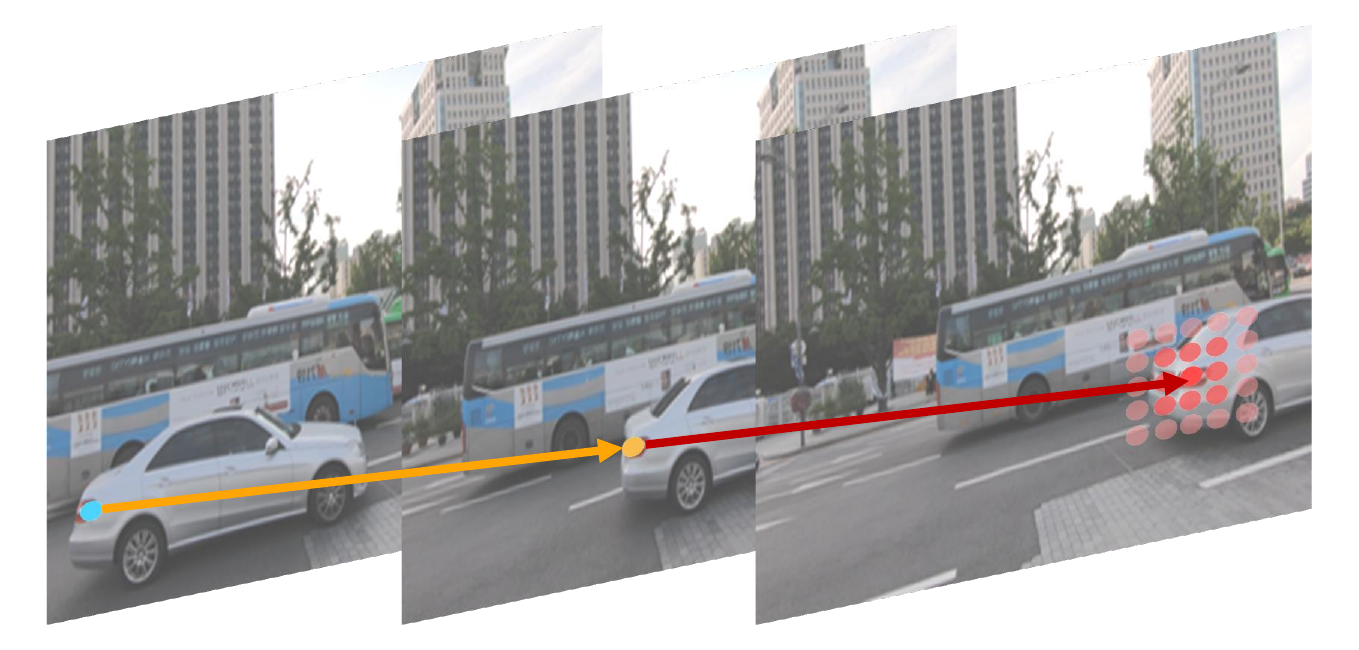}
\put(11,0.5){\small{Frame $t$-2}}
\put(38,0.5){\small{Frame $t$-1}}
\put(71,0.5){\small{Frame $t$}}
\end{overpic}}

\subfloat[CAM motion propagation]{\hspace{-0.4cm}
\begin{overpic}[width=1.06\linewidth]{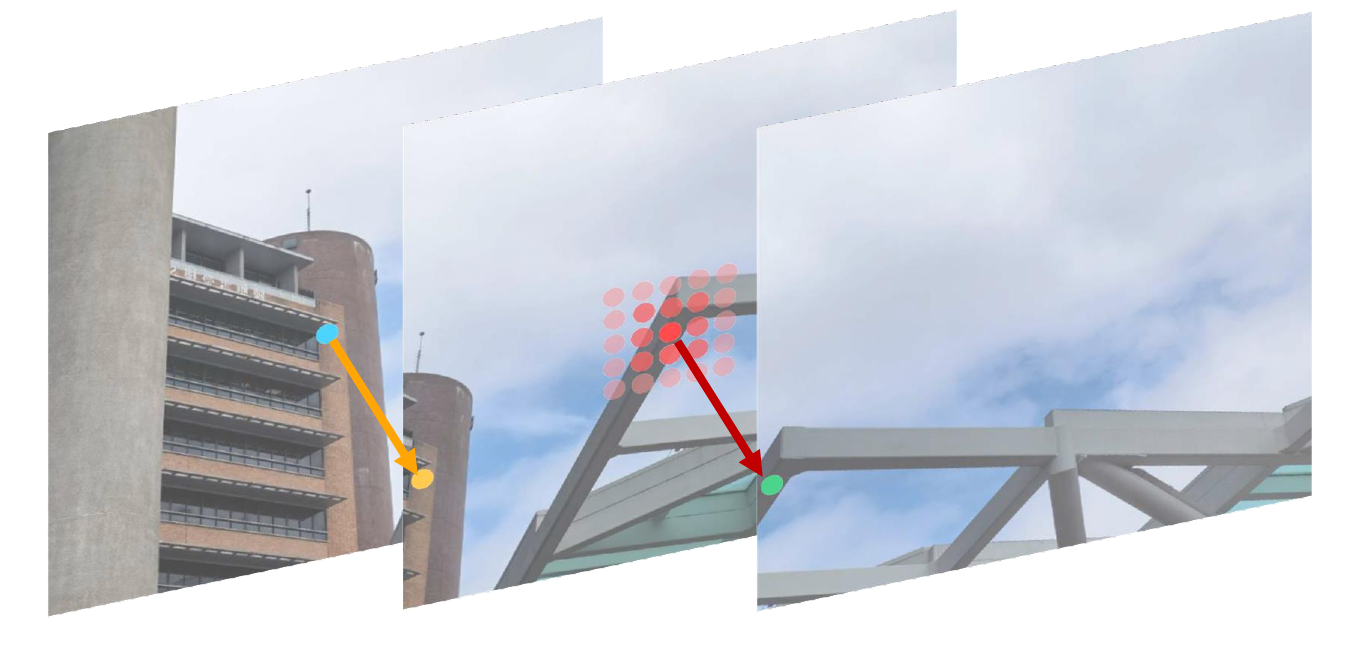}
\put(11,0.5){\small{Frame $t$-2}}
\put(38,0.5){\small{Frame $t$-1}}
\put(71,0.5){\small{Frame $t$}}
\end{overpic}}

\caption{Illustration of (a) object (OBJ) and (b) camera (CAM) motion propagation paths. The OBJ path aims to locate moving objects in the current frame, while the CAM path matches the static regions. The \textcolor{myorange}{orange arrow} represents the estimated motion from $I^{LR}_{t-2}$ to $I^{LR}_{t-1}$, which starts from the \textcolor{myblue}{blue point} and ends at the \textcolor{myorange}{orange point}. The \textcolor{myred}{red arrow} indicates the temporally propagated motion. 
In the CAM path, the \textcolor{mygreen}{green point} in $I^{LR}_{t}$ has the same position as the \textcolor{myorange}{orange point} in $I^{LR}_{t-1}$.
The \textcolor{myred}{red points} indicates the potential positions of the object at the corresponding frames, and the brighter colors represent higher likelihood.}
\label{fig:motions}
\end{figure}
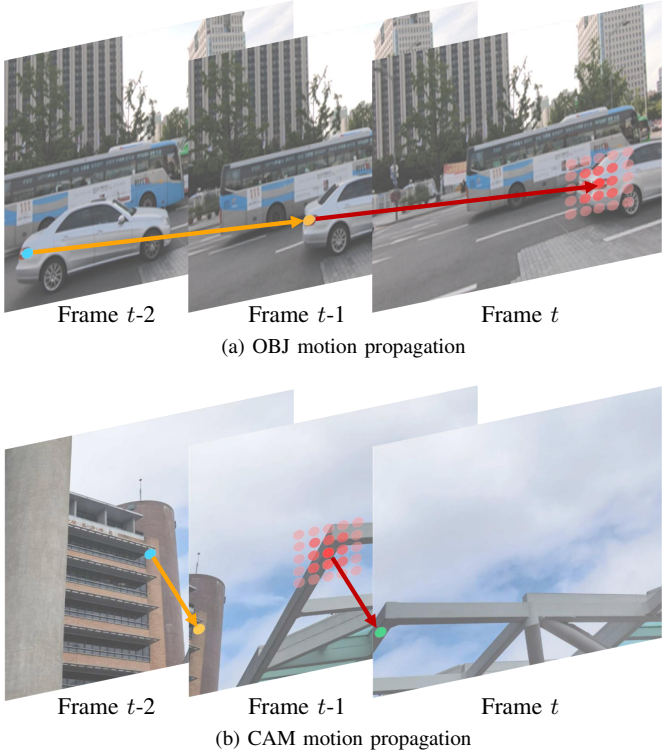

\textbf{OBJ Motion Propagation.}
There are two main types of motions between video frames.
The first type is for moving objects, termed as OBJ motion, which generally results from a combination of the movement of objects and the movement of  camera.
Based on the assumption that objects keep a relatively smooth movement between adjacent frames, we propagate the OBJ motion from the last frame to the current one.
As shown in Figure~\ref{fig:motions}(a), the object at the blue point $P^{i,j}_{t-2}$ in frame $I^{LR}_{t-2}$ moves to the orange point $P^{m,n}_{t-1}$ during the time interval $t-2$ to $t-1$.
By extrapolating this motion, the potential destination of this object should fall into the red regions in frame $I^{LR}_t$, where the brighter colors indicate candidates with higher likelihood.
Without losing generality, we assume that the potential destinations follow a two-dimensional Gaussian distribution centered at pixel $P^{2m-i, 2n-j}_t$.
Mathematically, the OBJ propagation path can be described as follows:
\begin{equation}
\label{eq1}
\begin{split}
&\begin{cases}
\boldsymbol{M}^{m,n}_{t-1} = (\Delta p_w, \Delta p_h),  \\
\boldsymbol{M}_t^{m-\Delta p'_w, n-\Delta p'_h} = (\Delta p'_w, \Delta p'_h), \\
{\mathbb{P}}(\Delta o_w, \Delta o_h) \sim \mathscr{N}(\boldsymbol{0}, \sigma \boldsymbol{I}), \\
\end{cases} \\
&\text{with}~\Delta o_w = \Delta p'_w - \Delta p_w, \Delta o_h = \Delta p'_h - \Delta p_h \\
\end{split}
\end{equation}

\noindent where $(\Delta p_w, \Delta p_h)$ denotes the motion inherited from the previous frame, and $(\Delta p'_w, \Delta p'_h)$ is the potential OBJ motion on current frame.
$\boldsymbol{M}^{m,n}_{t-1}$ describes the movement from pixel $P^{m+\Delta p_w,n+\Delta p_h}_{t-2}$ in frame $I^{LR}_{t-2}$ to pixel $P^{m,n}_{t-1}$ in frame $I^{LR}_{t-1}$. $\mathbb{P}$ represents the distribution of $\Delta o_w, \Delta o_h$, and $\sigma$ denotes the standard deviation of two-dimensional Gaussian distribution.
%
For each OBJ motion, we  select $k$ additional pixels as the candidates for offset estimation.

\textbf{CAM Motion Propagation.}
The other type of motion refers to the motions of static regions, which is actually caused by the movement of camera. Thus we call it as CAM motion.
Unlike OBJ motion, CAM motion has the characteristic that all static regions share the same camera motion, making it possible to spread the camera motion inherited from one region to other regions, especially for those unseen regions in the front frames.
Considering that the CAM motion is smooth among neighboring frames, we can transfer the CAM motion from the previous frame to the current one.
Figure~\ref{fig:motions}(b) shows an example of the CAM propagation path.
The region at blue point $P^{i,j}_{t-2}$ in frame $I^{LR}_{t-2}$ moves to the orange point $P^{m,n}_{t-1}$ in frame $I^{LR}_{t-1}$.
If both the green point $P^{m,n}_t$ and the orange point $P^{m,n}_{t-1}$ belong to static regions, we can easily infer that the motion of $P^{m,n}_t$ (red arrow) is similar to the one of $P^{m,n}_{t-1}$ (orange arrow).
Similar to the OBJ path, the CAM motion propagation path can be formulated as follows:
\begin{equation}
\label{eq2}
\begin{split}
&\begin{cases}
\boldsymbol{M}^{m,n}_{t-1} = (\Delta p_w, \Delta p_h),  \\
\boldsymbol{M}_t^{m, n} = (\Delta p'_w, \Delta p'_h), \\
{\mathbb{P}}(\Delta o_w, \Delta o_h) \sim \mathscr{N}(\boldsymbol{0}, \sigma\boldsymbol{I}), \\
\end{cases} \\
&\text{with}~\Delta o_w = \Delta p'_w - \Delta p_w, \Delta o_h = \Delta p'_h - \Delta p_h \\
\end{split}
\end{equation}
%
We also select $k$ candidates for each pixel in $I^{LR}_t$.

To simplify the implementation, we concurrently apply the two propagation paths to each pixel. This will not lead to conflicts since each path only generates a few candidates for the subsequent finetuning process.
\textbf{Finetuning.}
Since the movement of objects and cameras may slightly change in different frames, the motion fields propagated from past frames will inevitably contain some noise. To ensure that the propagated motion field fits the content of the current frame, we further refine it by a lightweight process with the contextual features $H^0_{t-1}$ and $H^0_t$.
The whole process is illustrated in the right part of Figure~\ref{fig:framework}.
During this process, the offset candidates of each pixel are first diffused to its neighbors to enhance the diversity and then evaluated iteratively to obtain the optimal candidate.
We introduce the following matching distance function to evaluate the offsets:
\begin{equation}
\label{eq3}
\begin{array}{l}
\text{d} = ||H^0_{t-1}(m+\Delta p'_w, n+\Delta p'_h) - H^0_t(m,n)||^2_2 .
\end{array}
\end{equation}
%

%
\subsection{Motion Confidence Weighted Fusion} \label{mcwf}
The alignment process will inevitably introduce errors in the motion field, which should be processed in the feature fusion process.
Most existing methods~\cite{vsrnet,edvr, iam} adopt a similarity-based re-weighting strategy to assign different importance to aligned features according to their similarities with reference features. 
However, in this way some well-matched features may be deemed as insignificant elements due to the low similarities caused by different yet complementary textures.
Unlike those methods, we propose to weigh the features based on the confidence of estimated offsets.

\begin{algorithm}[t]
\SetKwInput{INP}{Inp}
\SetKwInput{OUT}{Out}
\SetKwInput{RETURN}{Return}
\SetKwProg{PFOR}{PFor}{ Do}{End}
\SetKwProg{FOR}{For}{ Do}{End}

\caption{Temporal Motion Propagation Algorithm}\label{alg:alg1}

\INP{Features of frame $t$-1 $\boldsymbol{H}^0_{t-1}$; features of  frame $t$ $\boldsymbol{H}^0_t$; motion fields of frame $t$-1 $\boldsymbol{M}_{t-1}$; max iteration $k$; search range $\sigma$.}

\OUT{Motion fields of frame $t$ $\boldsymbol{M}_t$; motion confidences of frame $t$ $\boldsymbol{C}_t$\\}

\tcp*[h]{parallel for-loop in gpus}

\FOR{each $P^{m,n}_{t-1}$ in frame $t$-1 }{
Initialize candidates set $S_{m,n} = \{\}$;

\FOR{$i=0,1,\cdots,k-1$}{
Generate $(\Delta p'^{(i)}_w, \Delta p'^{(i)}_h)$ using Eq. \ref{eq1};

Add this candidate into $S_{m-\Delta p'^{(i)}_w, n-\Delta p'^{(i)}_h}$;

Generate $(\Delta p'^{(i)}_w, \Delta p'^{(i)}_h)$ using Eq. \ref{eq2};

Add this candidate into $S_{m, n}$.
}

Collect candidates from neighbors into $S_{m,n}$;

Compute distances for all candidates using Eq. \ref{eq3};

Update $\boldsymbol{M}^{m,n}_t$ based on the minimal distance;

Update $\boldsymbol{C}^{m,n}_t$ using Eq. \ref{eq4}.

}


\end{algorithm}
Specifically, as shown in the orange color box in Figure~\ref{fig:framework}, after extracting the visual features of the current frame $I^{LR}_t$, we introduce two independent branches for motion estimation and texture extraction, respectively.
The  former generates the motion features $H^0_{t}$, which are used to refine the motion field and evaluate the confidence of estimated offsets.
The latter extracts the texture features $H^1_{t}$ that are utilized to reconstruct the HR frame.
With such an architecture, we can assign larger weights to warped texture features containing complementary textures, even if they have low similarities to the reference texture features.
To further reduce the computations, we directly use the minimum distance map obtained in the aforementioned finetuning process to measure the precision of the estimated offsets.
The whole process of motion confidence weighted fusion (MCWF) can be described as follows:

\begin{equation}
\label{eq4}
\begin{cases}
\boldsymbol{C}_t = e^{-a\cdot D_{min}}, \\
\boldsymbol{\tilde H}^1_{t-1} = \boldsymbol{C}_t \cdot  \boldsymbol{\hat H}^1_{t-1},
\end{cases}
\end{equation}
where $\boldsymbol{\hat H}^1_{t-1}$ is the warped texture features from previous frame $I^{LR}_{t-1}$, $\boldsymbol{\tilde H}^1_{t-1}$ is the weighted texture features, $\boldsymbol{C}_t$ represents the confidence of refined motion filed $M_t$, and $D_{min}$ denotes the minimum distance map in the finetuning step of TMP.
Finally, the updated texture features $\boldsymbol{\tilde H}^1_{t-1}$ are concatenated with texture features of the current frame $H^1_t$, and then fed to the following reconstruction module. 

The pseudo-code of the proposed TMP algorithm is shown in \textbf{Algorithm \ref{alg:alg1}}.

\begin{table*}[!t]
\renewcommand{\arraystretch}{1.5}
\caption{
\begin{justify}
\justifying
Comparison with state-of-the-art online-VSR and non-online VSR methods. The runtime, frames per second (fps), params, and PSNR/SSIM metrics on three  benchmarks with two BI and BD degradations are reported.
The category (Cate.) and support frames (Supp.F.) of the competing methods are also indicated. 
``SW.'', ``Bi.'' and  ``Uni.'' denote the sliding-window based methods, bidirectional propagation methods and unidirectional propagation methods, respectively.
We use ``P'' to denote all previous frames, and ``P(n)'' to represent the $n$ preceding frames.
Similarly, ``F'' denotes all future frames while ``F(n)'' represents the $n$ forward frames.
%
%
%
%
Runtime is computed on an LR frame of size $180\times 320$.
Following DAP~\cite{dap}, those methods which can process 720p videos in at least 24fps are recognized as real-time (R-T.) methods.
We evaluate the complexity and speed (runtime/fps/MACs/Params) of all open-source online-VSR methods on the same setting.
%
%
%
%
%
The best and second best results among online-VSR methods on each test set are highlighted in \textbf{\textcolor{red}{red}} and \textbf{\textcolor{blue}{blue}}.
The symbol ``-'' means that the result is not available.
\end{justify} }
\label{tab:comparison}

 \renewcommand{\arraystretch}{1.5}
\scalebox{0.95}{

\begin{tabular}{l|l|c|c|c|c|c|c||c|c|c|c}
\hline
  \multirow{3}*{\rotatebox{90}{Cate.}} & \multirow{3}*{Method} & \multirow{3}*{Supp.F.} & \multirow{3}*{\rotatebox{90}{R-T.}} & \multirow{3}*{\makecell[c]{Run$\downarrow$\\(ms)} } & \multirow{3}*{\makecell[c]{fps$\uparrow$\\(1/s)} } & \multirow{3}*{\makecell[c]{MACs\\ (G)}} & \multirow{3}*{\makecell[c]{Params \\(M)}}& \multicolumn{2}{c|}{BI degradation}& \multicolumn{2}{c}{BD degradation} \\
 \cline{9-12}
   ~ & ~ & ~ & ~ & ~ & ~&~ &~ & \makecell[c]{\\[-0.2cm] REDS4(RGB)$\uparrow$\\ (PSNR/SSIM)} & \makecell[c]{\\[-0.2cm] Vid4(Y)$\uparrow$\\ (PSNR/SSIM)} & \makecell[c]{\\[-0.2cm] Vimeo-90K-T(Y)$\uparrow$\\ (PSNR/SSIM) } & \makecell[c]{\\[-0.2cm] Vid4(Y)$\uparrow$\\ (PSNR/SSIM) } \\
    
\hline\hline
 
 ~ & TOFlow~\cite{toflow} & P(3)+F(3) & \xmark & - & - & - & 1.4 & 27.98/0.7990 & 25.89/0.7651 & 34.62/0.9212 & - \\
 
 ~ & DUF~\cite{duf} & P(3)+F(3) & \xmark & 974 & 1.0 & - & 5.8 & 28.63/0.8251 & - & 36.87/0.9447 & 27.38/0.8329 \\
 
 ~ & PFNL~\cite{pfnl} & P(3)+F(3) & \xmark& 295 & 3.4 & - & 3.0 & 29.63/0.8502 & 26.73/0.8029 & - & 27.16/0.8355 \\
 
 \multirow{3}*{\rotatebox{90}{SW.}} & RBPN~\cite{rbpn} & P(3)+F(3) & \xmark &  1507 & 0.7 & - & 12.2 & 30.09/0.8590 & - & 37.20/0.9458 & -  \\

~ & MuCAN~\cite{mucan} & P(2)+F(2) & \xmark & - & - & $7'923$ & - & 30.88/0.8750 & - & - & - \\

~ & STAN~\cite{STAN} & P(10)+F(10) & \xmark & 108 & 9.30 & 980 & 16.2 & 28.85/0.8207 & 25.58/0.7430 & - & - \\

~ & TGA~\cite{tga} & P(3)+F(3) &\xmark& - & - & - & 5.8 & - & - & 37.59/0.9516 & 27.63/0.8423 \\

 ~ & EDVR-M~\cite{edvr} & P(2)+F(2) & \xmark & 118 & 8.5 & 462 & 3.3 & 30.53/0.8699 & 27.10/0.8186 & 37.33/0.9484 & 27.45/0.8406 \\
 
 ~ & EDVR~\cite{edvr} & P(2)+F(2) &\xmark& 378 & 2.6 & 2'017 & 20.6 & 31.09/0.8800 & 27.35/0.8264 & 37.81/0.9523 & 27.85/0.8503 \\

~ & IAM~\cite{iam} & P(3)+F(3) & \xmark & - & - & - & 17.0 & 31.30/0.8850 & 27.90/0.8380 & 37.84/0.9498 & - \\

~ & MCRNet~\cite{MCRNet} & P(3)+F(3) & \xmark & - & - & 910 & 14.0 & 31.86/0.9013 & 27.53/0.8397 & - & - \\

\hline\hline

 \multirow{3}*{\rotatebox{90}{Bi.}} & BasicVSR~\cite{basicvsr} & P+F & \xmark & 63 & 15.9 & 397 & 6.3 & 31.42/0.8909 & 27.24/0.8251 & 37.53/0.9498 & 27.96/0.8553 \\
 
~ & IconVSR~\cite{basicvsr} & P+F & \xmark & 70 & 14.3 & 452 & 8.7 & 31.67/0.8948 & 27.39/0.8279 & 37.84/0.9524 & 28.04/0.8570 \\

~ & BasicVSR++~\cite{basicvsr++} & P+F &\xmark & 77 & 13.0 & 418 & 7.3 & 32.39/0.9069 & 27.79/0.8400 & 38.21/0.9550 & 29.04/0.8753 \\

~ & SSL~\cite{SSL} & P+F &\xmark & 24 & 41.7 & 92 & 1.0 & 31.06/0.8933 & 27.15/0.8208 & 37.06/0.9458 & 27.56/0.8431 \\

\hline\hline

 \multirow{2}*{\rotatebox{90}{Uni.}} & RLSP~\cite{rlsp} & P+F(1) &\xmark & 49 & 20.4 & 252 & 4.2 & - & - & 36.49/0.9403 & 27.48/0.8388 \\


  
~ & ETDM~\cite{etdm} & P+F(3) &\xmark & 70 & 14.3 & - & 8.4 & 32.15/0.9024 & - & - & 28.81/0.8725 \\

\hline\hline
    
~ & Bicubic & N & \cmark & - & - & - & - & 26.14/0.7292 & 23.78/0.6347 & 31.30/0.8687 & 21.80/0.5246 \\
 
~ & FRVSR~\cite{frvsr} & P(1) & \xmark & 137 & 7.3 & - & 5.1 & - & - & 35.64/0.9319 & 26.69/0.8103  \\

 \multirow{4}*{\rotatebox{90}{Online-VSR}} & RSDN~\cite{rsdn} & P &\xmark & 49 & 20.4 & 356 & 6.2 & - & - & 37.23/0.9471 & \textbf{\textcolor{red}{27.92/0.8505}}  \\

~ & RRN~\cite{rrn} & P & \cmark & 34 & 29.4 & 193 & 3.4 & - & - & 36.69/0.9432 & \textbf{\textcolor{blue}{27.69/0.8488}}  \\


~ & BasicVSR++* & P & \cmark & 40 & 25.0 & 146 & 3.0 & 30.44/0.8686 & \textbf{\textcolor{blue}{27.06/0.8173}} & 37.11/0.9464 & 27.49/0.8426 \\ 

~ & DAP~\cite{dap} & P &\cmark & 38 & 26.3 & 165 & - & \textbf{\textcolor{blue}{30.59/0.8703}} & - & \textbf{\textcolor{blue}{37.29/0.9476}} & - \\ 

~ & CKBG~\cite{ckbg} & P &\cmark & \textbf{\textcolor{red}{9}} & \textbf{\textcolor{red}{111.1}} & 18 & 1.8 & 29.73/0.8514 & 26.34/0.7857 & - & - \\
\cline{2-12}
~ & \textbf{Ours} & P &\cmark & \textbf{\textcolor{blue}{25}} & \textbf{\textcolor{blue}{40.1}} & 176 & 3.1 & \textbf{\textcolor{red}{30.67/0.8710}} & \textbf{\textcolor{red}{27.10/0.8167}} & \textbf{\textcolor{red}{37.33/0.9481}} & 27.61/0.8428 \\\hline 
\end{tabular}}

\end{table*}

\section{EXPERIMENT} \label{sec:4}

\subsection{Experimental Settings}
\textbf{Datasets.}
We conduct experiments on three popular benchmarks, REDS~\cite{reds}, Vid4~\cite{vid4} and Vimeo-90K~\cite{toflow}. 
REDS is a high-quality video dataset which is widely used for evaluating the performance of VSR methods due to its various motions among frames.
Following EDVR~\cite{edvr}, we divide REDS into two parts.
The training set includes 266 clips (each has 100 consecutive frames), and the test set (REDS4) contains only four typical videos.
Vimeo-90K consists of about 90K 7-frame video clips.
The diverse scenarios in this dataset make it a popular benchmark for verifying the generalization performance of VSR models.
%
We train the models on the training set of Vimeo-90K and evaluate them on the corresponding test set (denoted by Vimeo-90K-T).
It should be noted that since Vimeo-90K-T is proposed for evaluating the performance of sliding-window VSR methods, we follow previous works~\cite{edvr, basicvsr++, dap} and only compute the PSNR and SSIM for the center frame of the 7-frames clip.
In addition, we directly utilize the models trained on Vimeo-90K to evaluate their performances on Vid4.

We adopt two different degradation strategies to perform $4\times$ downsampling, namely BI and BD.
The BI degradation refers to bicubic downsampling, which is implemented by using MATLAB function $\mathrm{imresize}$ in our experiments.
The BD refers to blur degradation.
We first blur the origin HR frame by a Gaussian filter with $\sigma=1.6$, and then conduct subsampling for every four pixels.
\textbf{Implementation Details.} 
We adopt 3 residual blocks (RB) to extract the features of each frame and another 30 RBs to reconstruct the HR frames and generate the hidden states. 
Two RBs and 2 conv layers are used for the motion branch, whose channel is reduced to 8 gradually. 
For the texture branch, a single conv layer is employed to extract features.
If not specified, the channel number is set to 64.

In the TMP module, $k$ is set to 2 while $\sigma$ is set to 30.
The $a$ in MCWF is set to $1$ for simplicity.
The downsampling factor is set to $s$ = 4 in our experiments.
During training, we feed the network with randomly cropped RGB sequences of size $64\times 64$ (low resolution) and adopt random flips, rotations, and temporal inversion to augment the data.
Similar to~\cite{edvr,basicvsr,basicvsr++}, the classical Charbonier loss~\cite{charbonierloss}, the Adam optimizer~\cite{adam}, and the Cosine Annealing scheme~\cite{cosine} are employed to optimize the network.
The initial learning rates for the motion branch and the rest of the network are set to $1e^{-3}$ and $1e^{-4}$, respectively.
The total number of iterations is 600K and the batch size is 8.
The model is trained under the PyTorch framework with two NVIDIA V100 GPUs.

When training on REDS~\cite{reds}, we set the batch size as 8, each of which contains 15 consecutive frames.
On Vimeo-90K~\cite{toflow}, since there are only 7 frames per clip, we temporally flip the original sequence to obtain a 14-frame sequence as input to the network.
During testing,  we evaluate the performance by taking the whole sequence as input.
%
%
%
%

%

\begin{figure*}[t]
\subfloat{}
\hfill
\begin{overpic}[width=\linewidth]{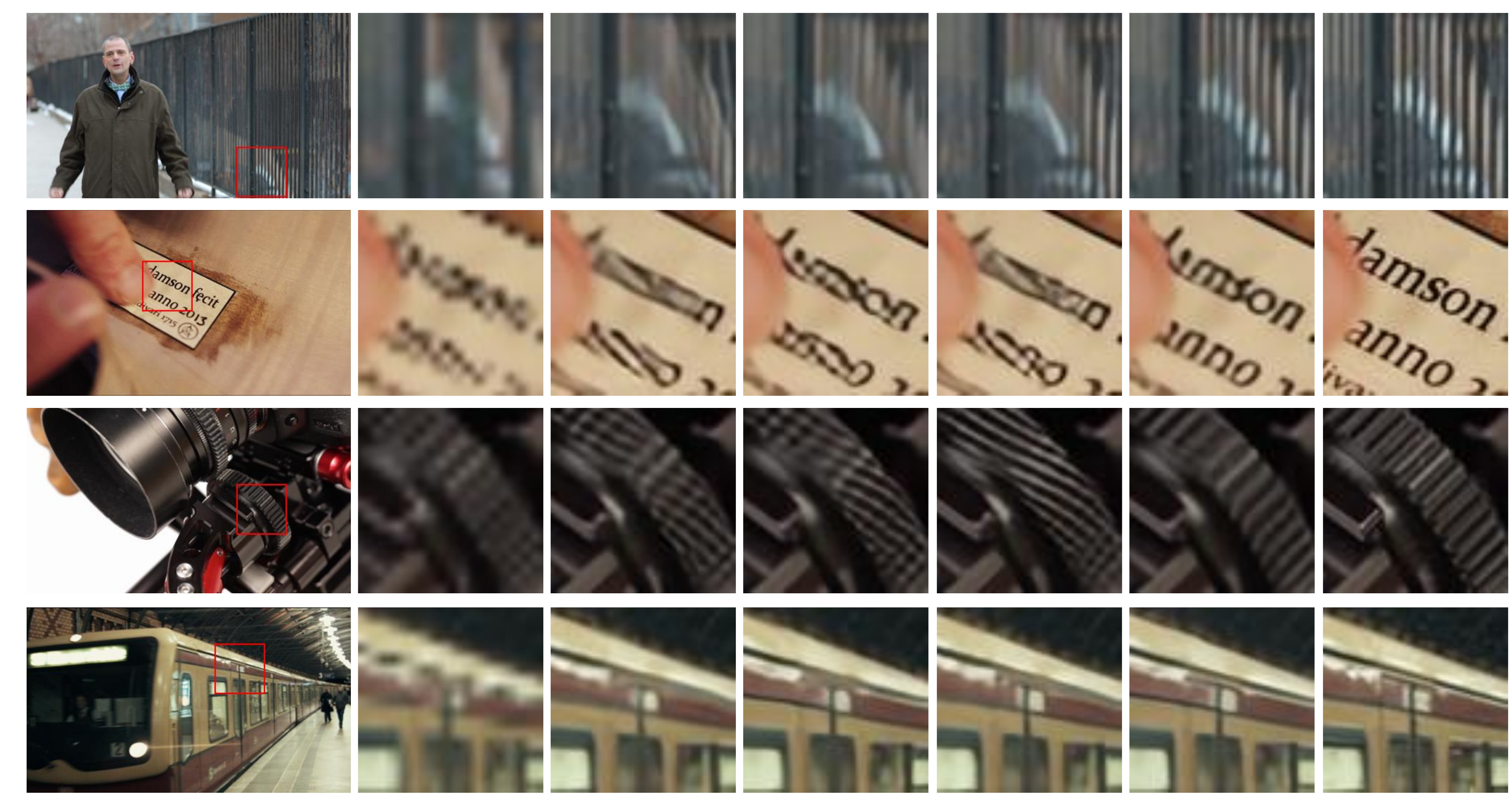}
\put(27, 53){\small{Bicubic}}
\put(40, 53){\small{RRN~\cite{rrn}}}
\put(50.5, 53){\small{BasicVSR++*}}
\put(64.5, 53){\small{RSDN~\cite{rsdn}}}
\put(79, 53){\small{\textbf{Ours}}}
\put(92.5, 53){\small{GT}}
\put(-0.2, 2.5){\rotatebox{90}{{Clip} 81/954}}
\put(-0.2, 16.3){\rotatebox{90}{{Clip} 3/670}}
\put(-0.2, 28.8){\rotatebox{90}{{Clip} 82/748}}
\put(-0.2, 41.8){\rotatebox{90}{{Clip} 24/315}}
\end{overpic}

\caption{Qualitative comparisons on details of static regions and moving objects.
The proposed method restores more details on both \textbf{static} regions (top two rows) and \textbf{moving} objects (bottom two rows) with CAM and OBJ motions, respectively. 
The results are from Vimeo-90K-T~\cite{toflow}.
Zoom in for the best view.}
\label{fig:experiment_fig1}

\end{figure*}

\begin{figure*}[!t]
\vspace{10pt}

\begin{minipage}{0.4\textwidth}
\begin{overpic}[width=\textwidth, height=0.6\textwidth]{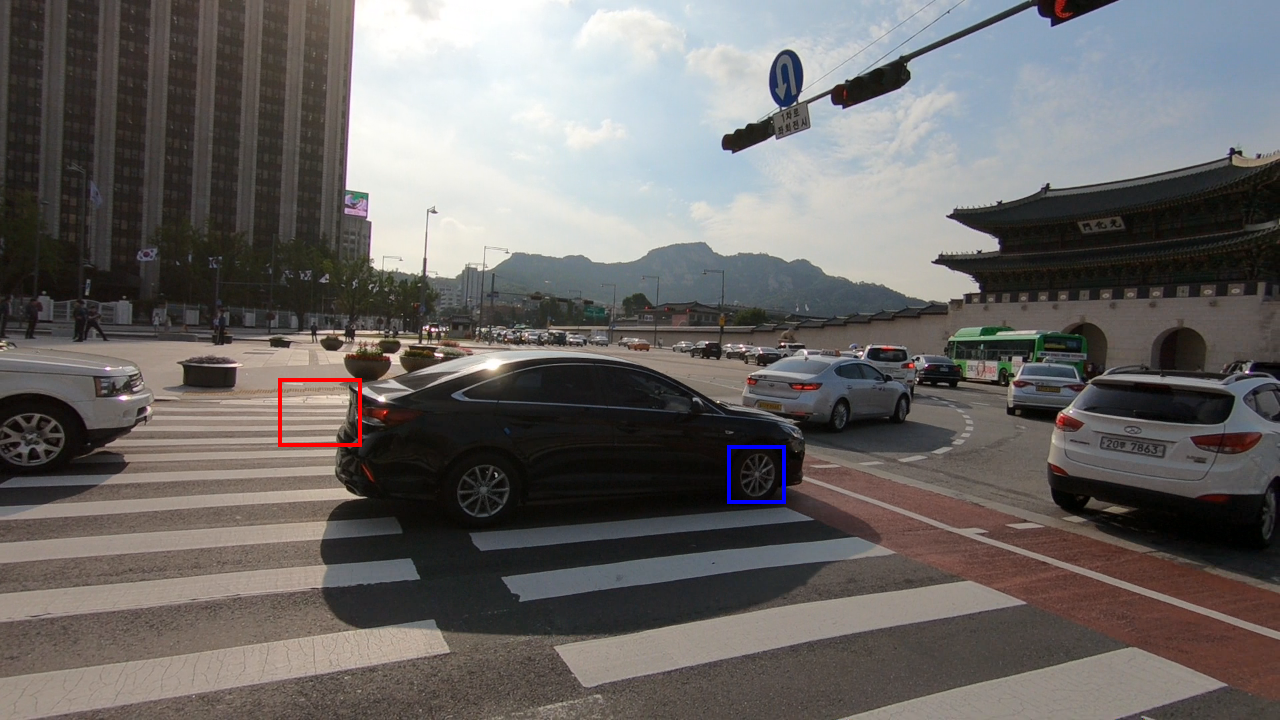}
\put(-5,20){\rotatebox{90}{Clip 015/095}}
\put(112, 60.5){\small{Bicubic}}
\put(140, 60.5){\small{CKBG~\cite{ckbg}}}
\put(168, 60.5){\footnotesize{BasicVSR++*}}
\put(204, 60.5){\small{\textbf{Ours}}}
\put(235.5, 60.5){\small{GT}}
\end{overpic}
\end{minipage}
\hspace{5pt}
\begin{minipage}{0.6\textwidth}
\includegraphics[width=0.19\textwidth, height=0.17\textwidth]{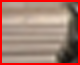}
\includegraphics[width=0.19\textwidth, height=0.17\textwidth]{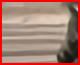}
\includegraphics[width=0.19\textwidth, height=0.17\textwidth]{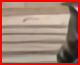}
\includegraphics[width=0.19\textwidth, height=0.17\textwidth]{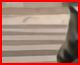}
\includegraphics[width=0.19\textwidth, height=0.17\textwidth]{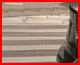}
\\[1em]
\includegraphics[width=0.19\textwidth, height=0.19\textwidth]{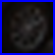}
\includegraphics[width=0.19\textwidth, height=0.19\textwidth]{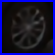}
\includegraphics[width=0.19\textwidth, height=0.19\textwidth]{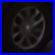}
\includegraphics[width=0.19\textwidth, height=0.19\textwidth]{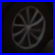}
\includegraphics[width=0.19\textwidth, height=0.19\textwidth]{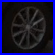}
\end{minipage}

\vspace{10pt}

\begin{minipage}{0.4\textwidth}
\begin{overpic}[width=\textwidth, height=0.6\textwidth]{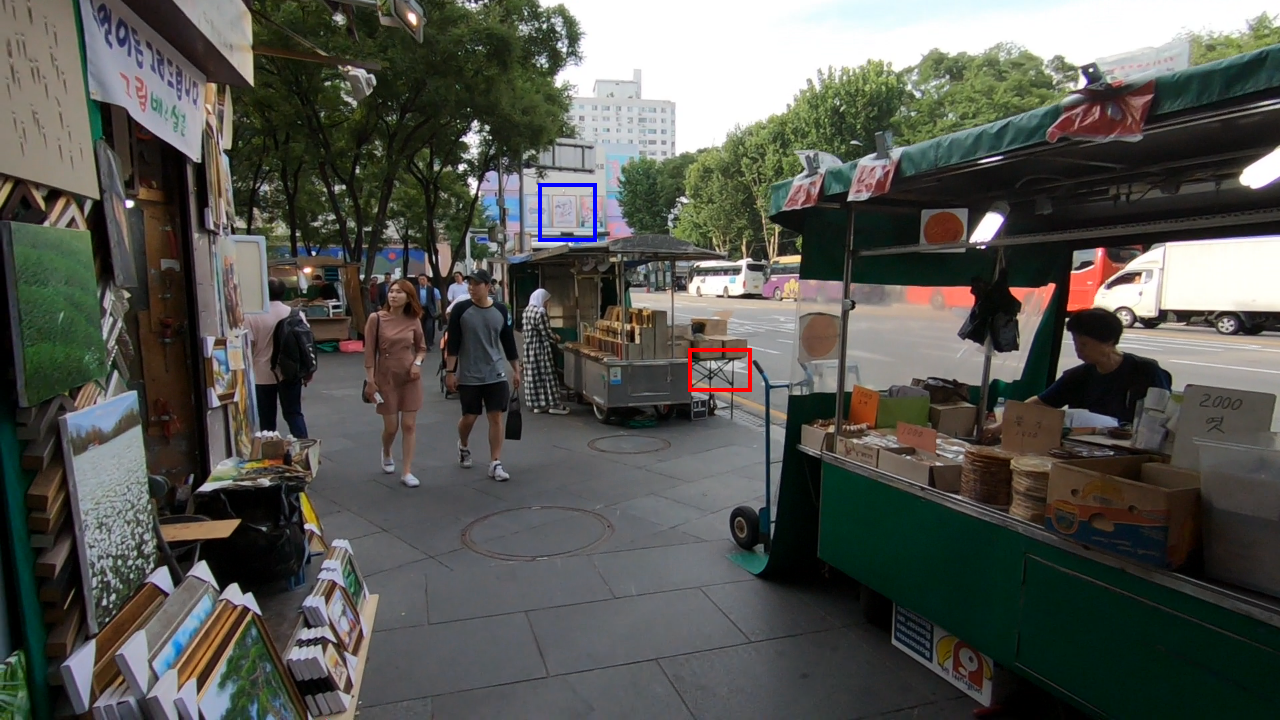}
\put(-5,20){\rotatebox{90}{Clip 020/008}}
\end{overpic}
\end{minipage}
\hspace{5pt}
\begin{minipage}{0.6\textwidth}
\includegraphics[width=0.19\textwidth, height=0.17\textwidth]{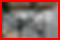}
\includegraphics[width=0.19\textwidth, height=0.17\textwidth]{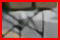}
\includegraphics[width=0.19\textwidth, height=0.17\textwidth]{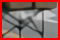}
\includegraphics[width=0.19\textwidth, height=0.17\textwidth]{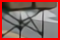}
\includegraphics[width=0.19\textwidth, height=0.17\textwidth]{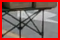}
\\[1em]
\includegraphics[width=0.19\textwidth, height=0.19\textwidth]{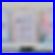}
\includegraphics[width=0.19\textwidth, height=0.19\textwidth]{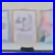}
\includegraphics[width=0.19\textwidth, height=0.19\textwidth]{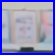}
\includegraphics[width=0.19\textwidth, height=0.19\textwidth]{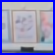}
\includegraphics[width=0.19\textwidth, height=0.19\textwidth]{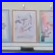}
\end{minipage}

\caption{Qualitative results on REDS4~\cite{reds}.
Our TMP method recovers more details on the frames that contain various motions. Zoom-in for best view.}
\label{fig:experiment_reds4}
\end{figure*}

\begin{figure*}[!t]
\centering
\begin{overpic}[width=\linewidth]{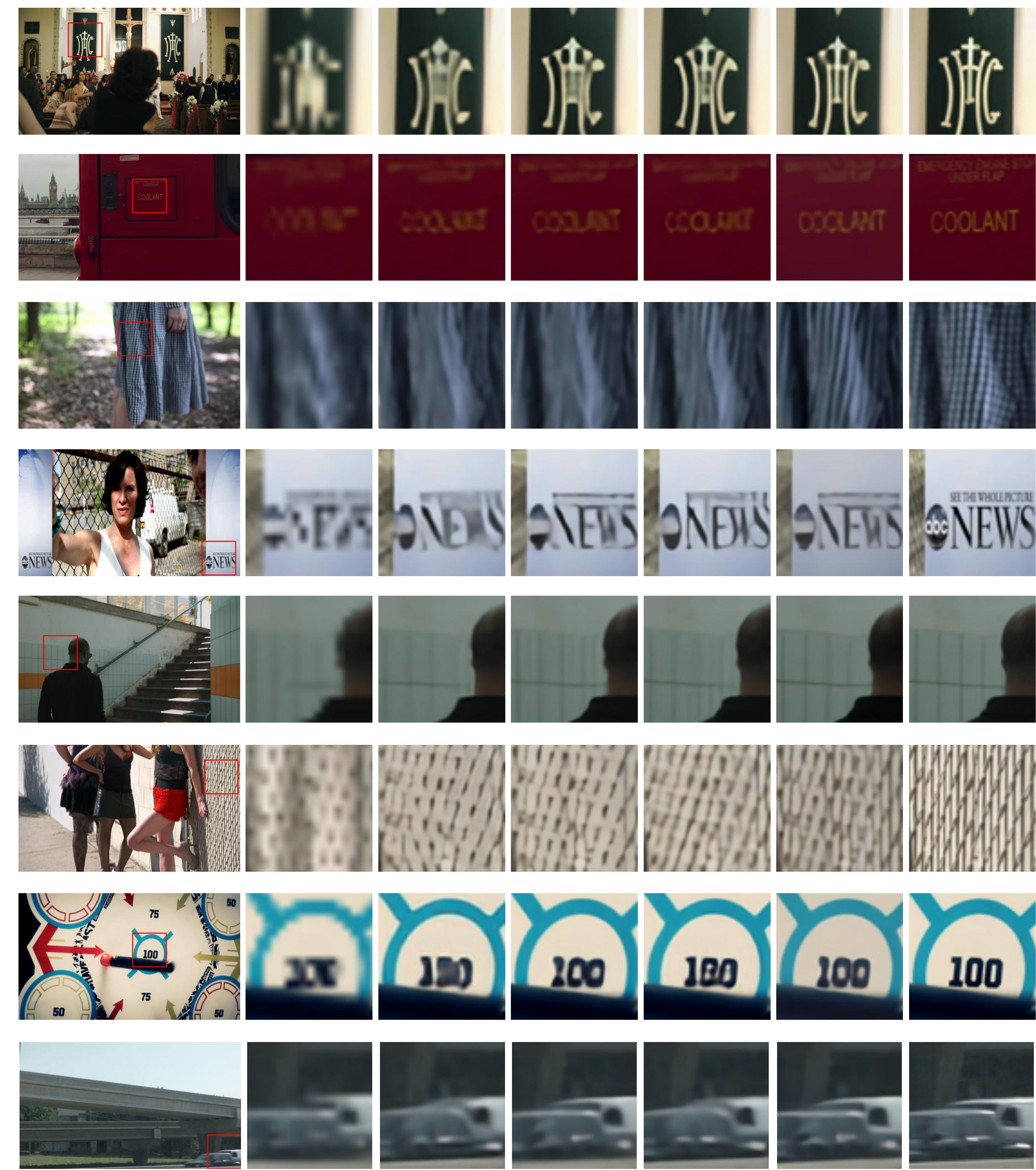}
\put(24, 99.5){\small{Bicubic}}
\put(34.5, 99.5){\small{RRN~\cite{rrn}}}
\put(44.6, 99.5){\small{BasicVSR++*}}
\put(57.4, 99.5){\small{RSDN~\cite{rsdn}}}
\put(70, 99.5){\small{\textbf{Ours}}}
\put(82, 99.5){\small{GT}}

\put(-0.2, 1.3){\rotatebox{90}{{Clip} 83/963}}
\put(-0.2, 14.1){\rotatebox{90}{{Clip} 80/837}}
\put(-0.2, 26.7){\rotatebox{90}{{Clip} 48/660}}
\put(-0.2, 39.7){\rotatebox{90}{{Clip} 34/986}}

\put(-0.2, 52.2){\rotatebox{90}{{Clip} 25/898}}
\put(-0.2, 64.8){\rotatebox{90}{{Clip} 12/040}}
\put(-0.2, 77.5){\rotatebox{90}{{Clip} 10/935}}
\put(-0.2, 90.6){\rotatebox{90}{{Clip} 1/837}}
\end{overpic}

\caption{More qualitative results on Vimeo-90K-T~\cite{toflow}.
The proposed TMP method generates better textures on various scenes. 
Zoom-in for best view.}
\label{fig:experiment_fig1_sup}
\end{figure*}

\begin{figure*}[!t]
\vspace{10pt}
\begin{overpic}[width=0.49\textwidth, height=0.30\textwidth]{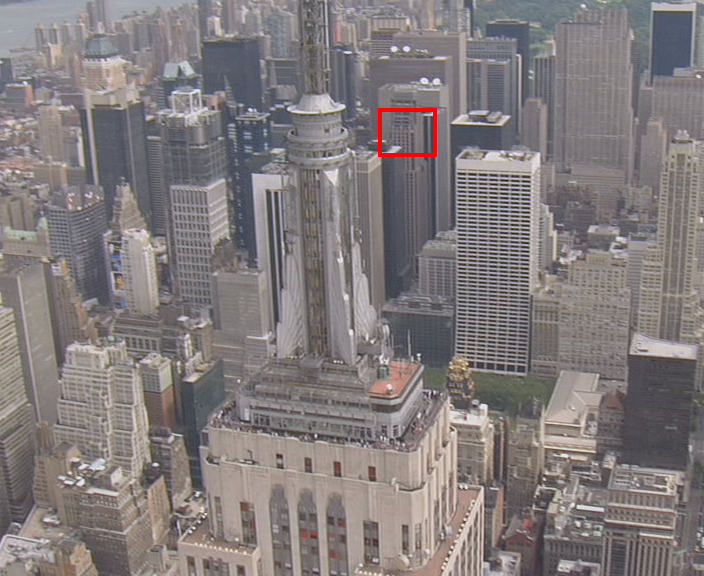}
\put(30,-5){The $29$-th frame of City clip}
\put(130,-5){The $7$-th frame of Walk clip}
\put(9, -58){Bicubic}
\put(33, -58){FRVSR~\cite{frvsr}}
\put(64, -58){RRN~\cite{rrn}}
\put(90, -58){BasicVSR++*}
\put(120.5, -58){RSDN~\cite{rsdn}}
\put(153.5, -58){\textbf{Ours}}
\put(183.5, -58){GT}
\end{overpic}
\includegraphics[width=0.49\textwidth, height=0.30\textwidth]{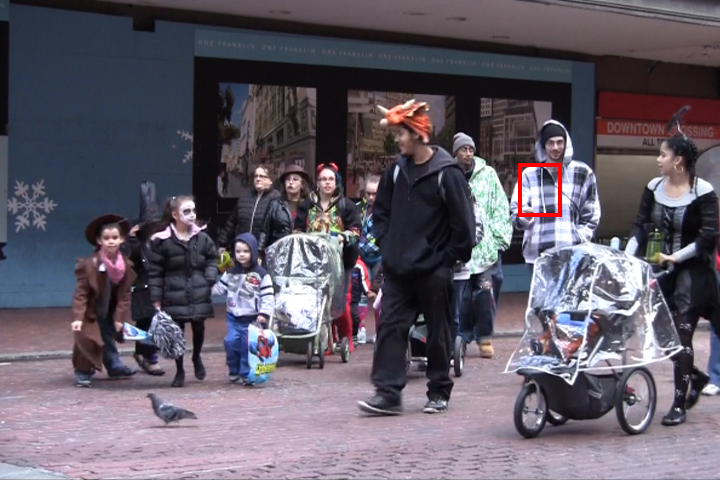}
\\[2em]
\includegraphics[width=0.135\textwidth]{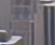}
\includegraphics[width=0.135\textwidth]{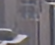}
\includegraphics[width=0.135\textwidth]{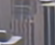}
\includegraphics[width=0.135\textwidth]{figures/vid4/basicvsr.p2.city.png}
\includegraphics[width=0.135\textwidth]{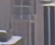}
\includegraphics[width=0.135\textwidth]{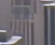}
\includegraphics[width=0.135\textwidth]{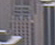}
\\[0.3em]
\includegraphics[width=0.135\textwidth,height=0.11\textwidth]{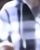}
\includegraphics[width=0.135\textwidth,height=0.11\textwidth]{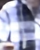}
\includegraphics[width=0.135\textwidth,height=0.11\textwidth]{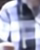}
\includegraphics[width=0.135\textwidth,height=0.11\textwidth]{figures/vid4/basicvsr.p2.walk.png}
\includegraphics[width=0.135\textwidth,height=0.11\textwidth]{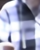}
\includegraphics[width=0.135\textwidth,height=0.11\textwidth]{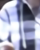}
\includegraphics[width=0.135\textwidth,height=0.11\textwidth]{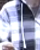}

\vspace{10pt}

\caption{Qualitative results on Vid4~\cite{vid4}.
%
%
The proposed TMP method obtains better visual results . Zoom-in for best view.}
\label{fig:experiment_vid4}
\end{figure*}

\begin{figure}[t]
\subfloat{}
\hfill
\scalebox{0.95}{
\begin{overpic}[width=\linewidth]{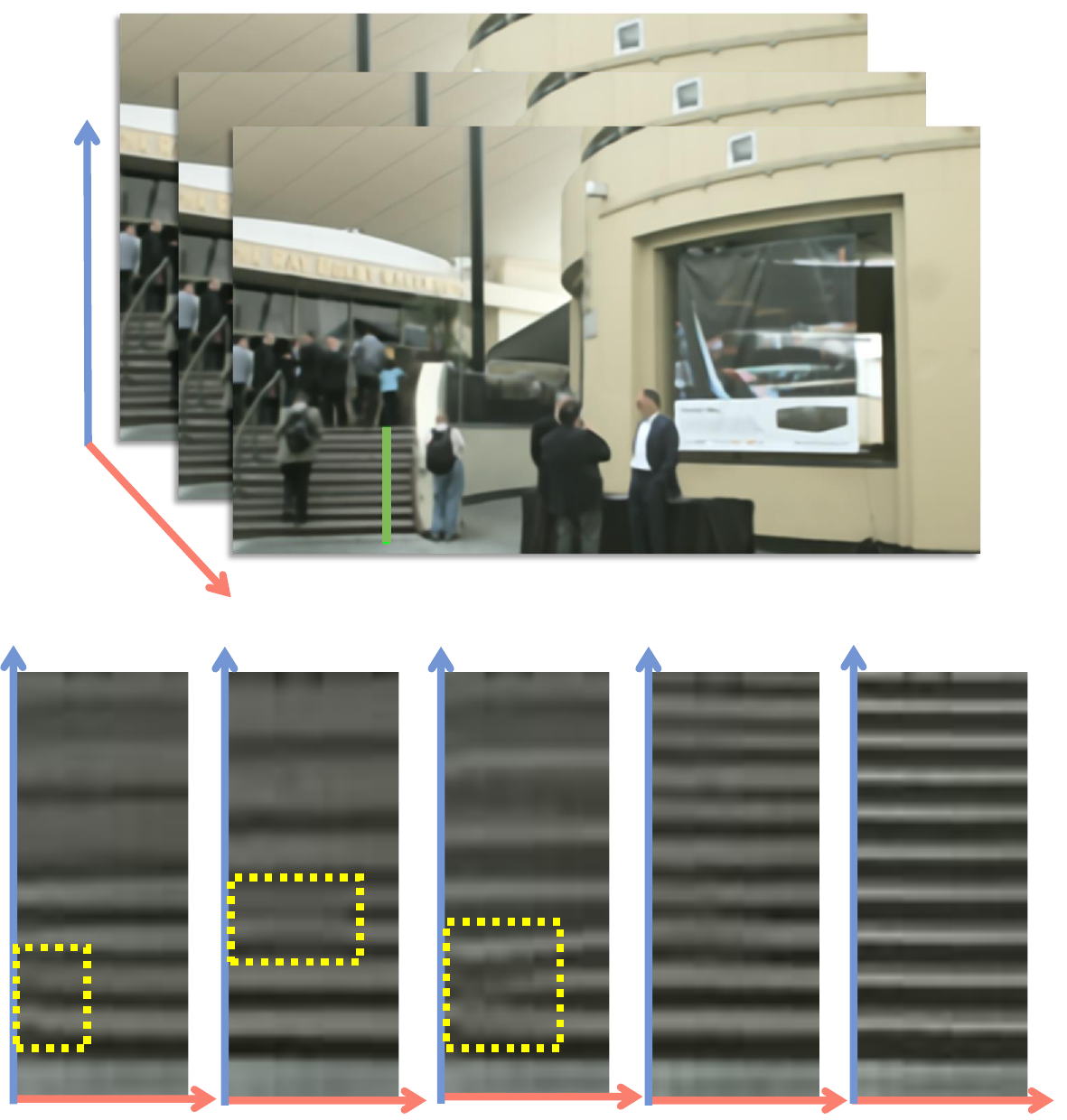}
\put(3.8, 78){\Large{$y$}}
\put(12.3,49.3){\Large{$t$}}
\put(7, -2){\small{(a)}}
\put(26, -2){\small{(b)}}
\put(44.5, -2){\small{(c)}}
\put(63, -2){\small{(d)}}
\put(82, -2){\small{(e)}}
\end{overpic}
}

\caption{
Visualization of Temporal Consistency.
From left to right: the temporal consistency profile generated by (a) RRN~\cite{rrn}, (b) BasicVSR++*, (c) RSDN~\cite{rsdn}, (d) ours and (e) groundtruth.
We observe the changes of pixels on the green line across time.
The camera is moving horizontally.
The results of other methods all contain irregular patterns, as indicated by the yellow dashed box, while our method keeps the pattern well.
}
\label{fig:experiment_fig2}
\end{figure}
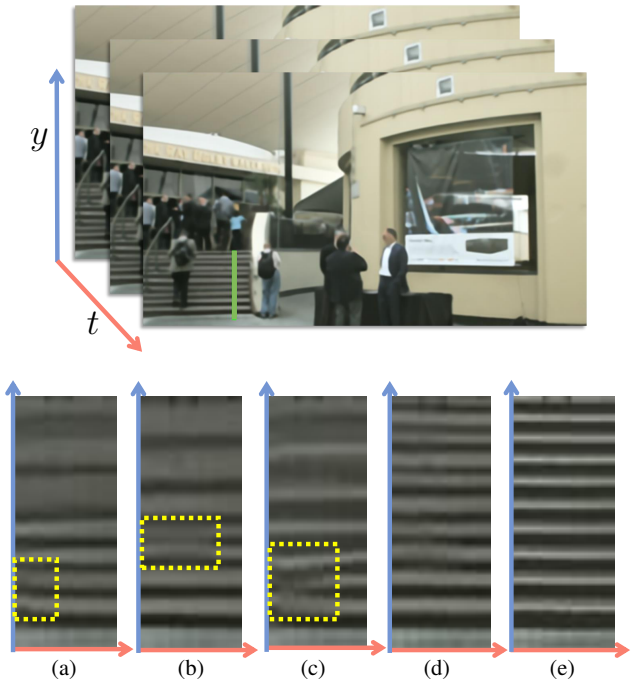

\textbf{Compared Methods.}
We compare our method with state-of-the-art online-VSR methods (FRVSR~\cite{frvsr}, RSDN~\cite{rsdn}, RRN~\cite{rrn}, DAP~\cite{dap} and CKBG~\cite{ckbg}), as well as sliding-window based methods (TOFlow~\cite{toflow}, DUF~\cite{duf}, PFNL~\cite{pfnl}, RBPN~\cite{rbpn}, MuCAN~\cite{mucan}, TGA~\cite{tga}, EDVR~\cite{edvr}, IAM~\cite{iam} and STAN~\cite{STAN} and MCRNet~\cite{MCRNet}), bidirectional propagation methods (BasicVSR~\cite{basicvsr}, IconVSR~\cite{basicvsr},  BasicVSR++~\cite{basicvsr++} and SSL~\cite{SSL}), and unidirectional propagation methods (RLSP~\cite{rlsp} and ETDM~\cite{etdm}). 
%
Additionally, we introduce a method ``BasicVSR++*'' by removing backward propagation branches of BasicVSR++ and reducing its model size to meet the access and latency constraints of online-VSR applications.

\subsection{Results}  \label{comparision}
\textbf{Comparison with Online-VSR Methods.} The quantitative results are listed in the bottom of Table~\ref{tab:comparison}. One can see that our method significantly surpasses other online models in terms of balanced accuracy and speed. 
Though CKBG has the fastest speed, it has poor PSNR/SSIM performance. Our method outperforms CKBG by 0.94dB and 0.76dB on REDS4 and Vid4, respectively.
Compared to DAP, our method achieves a performance gain of 0.08dB on the RESD4 dataset and it is 1.52 times faster (25ms vs. 38ms).
Our method also exhibits superior performance (37.33dB in PSNR and 0.9481 in SSIM) on the large-scale Vimeo-90K-T dataset, which demonstrates its capacity to effectively handle various scenarios.
%
The PSNR/SSIM indices of our method are slightly lower than that of RSDN on BD-degrated Vid4 dataset.
This is because the complex textures of Vid4 are heavily corrupted by the BD blur degradation, while RSDN employs much more parameters to solve this issue, making it much slower.

%
%
Figure~\ref{fig:experiment_fig1} present qualitative comparisons on the recovered details of \textbf{static} regions and \textbf{moving} objects, respectively.  
One can see that our method shows better visual quality than other online-VSR methods on both scenes.
This can be attributed to the inherited motion fileds through CAM and OBJ motion propagation during inference.
For example, the image in the first row comes from the ``00081/0954'' clip, where the camera is  moving slowly.
Our method recovers more iron bars of the barrier and shows clearer details. The image displayed in the third row is from the ``00082/0748'' clip, where the gear exhibits rotational movement.
Other online-VSR methods fail to accurately align the frames and result in wrong structures of the gear, while our method precisely tracks the rotation and successfully recover the ridges.

%
Figure~\ref{fig:experiment_reds4}
, Figure~\ref{fig:experiment_fig1_sup} and Figure~\ref{fig:experiment_vid4} illustrate more visual results on REDS4~\cite{reds}, Vimeo-90K~\cite{toflow} and Vid4~\cite{vid4}, respectively.
Although videos from REDS4 contain relatively larger motions, one can see from Figure~\ref{fig:experiment_reds4} that our method still recovers better details than competing methods.
Vimeo-90K contains various scenes with different objects. 
%
%
As can be seen from Figure~\ref{fig:experiment_fig1_sup}, the enriched textures in the VSR results further affirm the effectiveness of our TMP approach.
The images from Vid4 dataset are heavily corrupted by BD downsampling, making it harder for restoration.
However, our TMP method still reconstructs compelling details, as demonstrated in Figure~\ref{fig:experiment_vid4}.
%


\textbf{Comparison with Non-Online VSR methods.} We further compare the results of our method with non-online VSR methods. 
As shown in Table~\ref{tab:comparison}, there exist two key factors that affect the accuracy of online-VSR and non-online VSR methods.
First, the set of support frames heavily affects the PSNR/SSIM results because it determines the amount of information that a VSR model can exploit.
BasicVSR++, which belongs to the bidirectional propagation methods, aggregates information from all video frames and achieves the best PSNR/SSIM indices on all the three benchmarks.
ETDM accesses more previous frames and obtains an improvement of about 0.85dB on REDS4 over IAM.
Second, the network capacity plays an important role in the performance of VSR methods.
We see that most online-VSR methods have half the parameters of the state-of-the-art non-online VSR methods (\eg, 3.1M for our model while 8.4M for ETDM, 7.3M for BasicVSR++, and 17.0M for IAM).

However, the longer latency of non-online VSR methods hinders them from the online-VSR applications. 
For a fair comparison, we further train a variant of BasicVSR++, namely BasicVSR++*, which only accesses the previous video frames and has a comparable number of parameters with our model. As can be seen from Table~\ref{tab:comparison}, BasicVSR++* performs much worse than our model in terms of either PSNR/SSIM or fps indices, validating the effectiveness and efficiency of the proposed TMP and MCWF strategies.

\textbf{Complexity and Speed.} 
From Table~\ref{tab:comparison}, one can see that our method achieves the second fastest inference speed (25ms/40.1fps) among all VSR methods with relatively good PSNR/SSIM performance.
For online-VSR methods, DAP, BasicVSR++* and RRN consume nearly the same amount of MACs as our method, but they have obviously longer latency (38ms, 40ms and 34ms, respectively).
On the other hand, EDVR-M has a similar PSNR/SSIM performance to our method, but its inference speed is about five times slower (fps: 8 vs. 40).
PFNL contains a similar number of parameters to ours, yet it takes nearly 300ms to process a frame of $180\times 320$ resolution.
In summary, compared with those methods having similar complexity, parameters or performance, our method runs much faster.
\begin{table}
\centering
 \caption{Temporal consistency on Vimeo-90K.}
 \renewcommand{\arraystretch}{1.6}
\scalebox{0.95}{
\begin{tabular}{l|@{\hspace{3pt}}c@{\hspace{3pt}}|@{\hspace{3pt}}c@{\hspace{3pt}}|@{\hspace{3pt}}c@{\hspace{3pt}}|@{\hspace{3pt}}c@{\hspace{3pt}}|@{\hspace{3pt}}c@{\hspace{3pt}}}
\hline 
Metrics & Bicubic & RRN~\cite{rrn} & RSDN~\cite{rsdn} & BasicVSR++* & \textbf{TMP} \\ 
\hline \hline
tOF$\downarrow$ \scalebox{0.7}{$\times  10$} & 6.43 & 2.23  &  2.21 & 1.98 & 2.02 \\ \hline
TCC$\uparrow$ \scalebox{0.7}{$ \times  10$}  & 2.36 & 3.40  &  3.64 &  3.44 & 3.53 \\ \hline
FWE$\downarrow$ \scalebox{0.7}{$ \times 10^4 $} & -- &  3.32 & 3.45 &  3.07 & 3.14 \\ \hline
\end{tabular}}
 \label{tab:tc}
\end{table}

\textbf{Temporal Consistency.} 
%
We use three metrics tOF~\cite{tOF}, TCC~\cite{TCC} and Flow Warping Error (FWE)~\cite{FWE} to measure the temporal consistency. The comparison results are presented in Table~\ref{tab:tc}.
It can be seen that our method achieves better temporal consistency than most methods. Though BasicVSR++* slightly outperforms our method, it adopts a sophisticated and cumbersome alignment module which lags behind us in efficiency (please refer to Table~\ref{tab:comparison}).
%
%
We also provide the visualization results in Figure~\ref{fig:experiment_fig2}.
Taking the clip ``00023/0216'' of Vimeo-90K-T as example, where the camera moves horizontally, our method produces less distortions and generates a more consistent profile than the competing methods across frames. This further demonstrates the effectiveness of our TMP module in frame alignment.

\begin{table}

\caption{
\begin{justify}
\justifying
Ablation Studies on TMP and MCWF. Evaluations are conducted on the REDS4 dataset and the fps is measured with the HR frame of size $1280\times 720$.
\end{justify}
}
\label{tab:ablation_1}

\centering
\renewcommand{\arraystretch}{1.5}
\begin{tabular}{c|p{0.3cm}p{0.3cm}p{0.3cm}p{0.3cm}p{0.7cm}|c|c}
\hline
 Model \# & Rand & OBJ & TMP & SRF & MCWF & fps(1/s) & PSNR/SSIM\\
\hline
1 & ~ & ~ & ~ & ~ & ~ & 40.52 & 29.74/0.8507 \\ 
\hline
2 & \;\;\checkmark & ~ & ~ & ~ & ~ & 40.45 & 29.92/0.8541 \\ 
3 & ~ & \;\;\checkmark & ~ & ~ & ~ & 40.31 & 30.33/0.8649 \\ 
4 & ~ & ~ & \;\;\checkmark & ~ & ~ & 40.30 & 30.41/0.8651 \\ 
5 & ~ & ~ & \;\;\checkmark & \;\;\;\checkmark & ~ & 40.09 & 30.46/0.8666  \\ 
6 & ~ & ~ & \;\;\checkmark & ~ & \;\;\;\checkmark & 40.13 & 30.60/0.8694 \\ 
\hline
\end{tabular}

\end{table}

\subsection{Ablation Studies}
In this section, we perform several ablation studies on the proposed TMP module and MCWF strategy to better illustrate their roles. 
We use REDS for training and REDS4 for evaluation, and set the number of iterations to 300K.

\textbf{Effectiveness of TMP and MCWF.} \label{AblaionStudies}
To validate the effectiveness of our proposed TMP module and MCWF strategy, we train five variant models of our method. 
As shown in Table~\ref{tab:ablation_1}, the baseline model (Model 1) directly fuses the hidden states of the last frame without alignment, which shows poor performance.
Model 2 performs alignment with a warp operation.
Different from Model 4 that adopts the proposed TMP module, Model 2 randomly initializes the motion field and then refines it using the same fintuning process as the TMP module.
Model 2 achieves a $0.18$dB improvement over Model 1, while lagging $0.49$dB behind Model 4.
The performance gap demonstrates the effectiveness of TMP in inheriting motion field from the previous frame.
Besides, TMP has minimal impact on the inference speed, only slightly dropping from 40.52fps to 40.30fps.
Model 3 utilizes only the OBJ path to inherit the offsets, while Model 4 applies both the OBJ and CAM paths.
One can see that CAM can bring additional improvement to OBJ since the CAM path can be applied to a broader range of
regions, such as unseen regions in previous frames. 
Model 5 employs the similarity reweighting fusion (SRF) strategy, which assigns importance to the warped features according to their similarities with the reference features.
Model 6 is the proposed network which applies the TMP module and MCWF strategy simultaneously.
One can see that our MCWF strategy achieves 0.27dB and 0.14dB performance gains over Model 4 and Model 5, respectively.

\begin{table}

\caption{
\begin{justify}
\justifying
Hyper-parameter (HP) Selection of TMP. Evaluations (PSNR/ SSIM) are conducted on REDS4 in two different color spaces. ``Y'' represents YCrCb color space, while ``RGB'' denotes the normal RGB color space.
\end{justify}
}
\label{tab:ablation_2}
 \renewcommand{\arraystretch}{1.5}
\centering
\begin{tabular}{>{\centering\arraybackslash}p{1.3cm}|>{\centering\arraybackslash}p{1.3cm}|>{\centering\arraybackslash}p{2cm}|>{\centering\arraybackslash}p{2cm}}
\hline
 HP & Values & REDS4(Y) & REDS4(RGB)\\
\hline
\hline
\multirow{4}*{k} & 0 & 31.72/0.8786 & 30.35/0.8635 \\
~ & 1 & 31.95/0.8838 & 30.60/0.8693 \\
~ & 2 & 31.96/0.8839 & 30.60/0.8694 \\
~ & 4 & 31.96/0.8839 & 30.60/0.8694 \\
\hline
\multirow{5}*{$\sigma$} & 0.1 & 31.72/0.8788 & 30.36/0.8637 \\
~ & 1 & 31.92/0.8828 & 30.55/0.8682 \\
~ & 10 & 31.96/0.8838 & 30.59/0.8693 \\
~ & 30 & 31.96/0.8839 & 30.60/0.8694 \\
~ & 60 & 31.76/0.8796 & 30.40/0.8648 \\
\hline
\end{tabular}

\end{table}

\textbf{Hyper-parameter Selection of TMP.} 
There are two major hyper-parameters in our TMP module.
The first one is $k$, which counts the offset candidates generated by each propagation path.
The second one is the standard deviation $\sigma$ of the Gaussian distribution, which implicitly impacts the searching range of estimated motions.

Table~\ref{tab:ablation_2} shows the ablation study results on the two hyper-parameters.
One can see that there is a significant performance decrease (about 0.24dB and 0.25dB on Y and RGB color spaces, respectively) when $k$ is set to 0, which indicates that directly using the inherited offsets is not the best choice for the alignment of current frame.
It is important to generate a few new offset candidates for further evaluation.
The best results are achieved when $k$ is set to 2. That is, introducing only a few more candidates will be enough for TMP to estimate the offsets.
The performance saturates when $k$ is further increased.

Regarding $\sigma$, the best results ($31.96$dB and $30.60$dB on Y and RGB color spaces, respectively) are achieved when it is set between $10$ and $30$.
A smaller $\sigma$ limits the search space, which decreases the accuracy of offset estimation for objects that change their motion.
A larger $\sigma$ represents a wider range of searching space, which however lowers the likelihood of the optimal candidates.

\begin{figure}[t]

\centering
\subfloat[]{\includegraphics[width=0.24\textwidth]{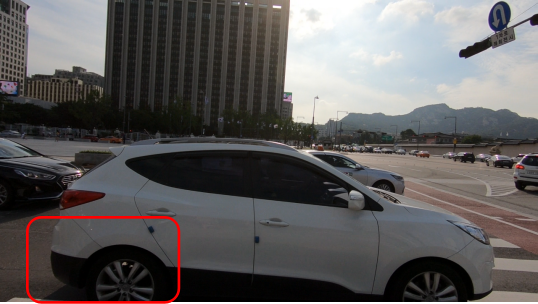}}
\hfill
\subfloat[]{\includegraphics[width=0.24\textwidth]{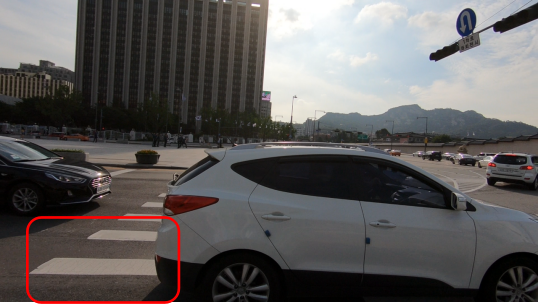}}\\
\vspace{-7pt}
\subfloat[]{\includegraphics[width=0.24\textwidth]{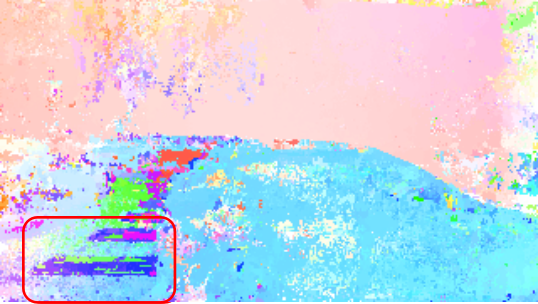}}
\hfill
\subfloat[]{\includegraphics[width=0.24\textwidth]{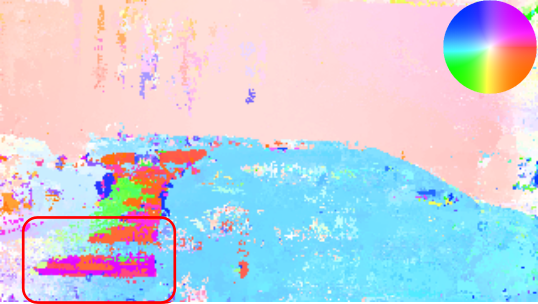}}\\
\vspace{-7pt}
\subfloat[]{\includegraphics[width=0.24\textwidth]{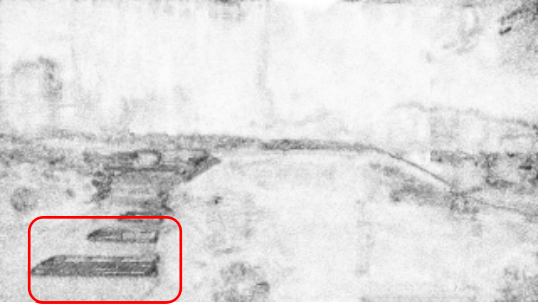}}
\hfill
\subfloat[]{\includegraphics[width=0.24\textwidth]{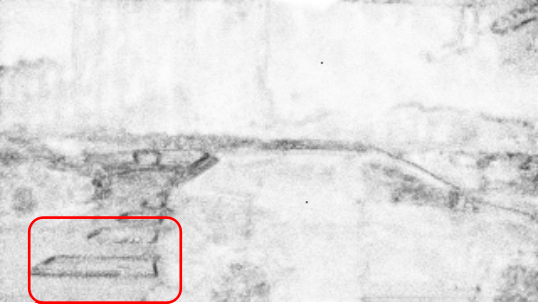}}

\caption{Visualizations of OBJ \& CAM paths and motion confidence. (a) The previous frame (T-1). (b) The reference frame T. (c) Offset field with only OBJ path. (d) Offset field with both paths. (e) Motion confidence with only OBJ path. (f) Motion confidence with both paths.} \label{fig:visual_result}
\end{figure}

\textbf{Visulization of Motion Field and Confidence.} The visualizations are shown in Figure~\ref{fig:visual_result}, where the cars move to the right while the buildings/roads shift to the left relative to the camera.
The OBJ path can only handle the motion of objects that appear in consecutive frames but cannot track the motion of newly appeared objects.
As can be seen in Figure~\ref{fig:visual_result}(c), the offsets of car and building can be accurately estimated with only the OBJ path since they appear in the two consecutive frames in Figure~\ref{fig:visual_result}(a) and Figure~\ref{fig:visual_result}(b). However, the motion of the crosswalk that does not appear in the previous frame in Figure~\ref{fig:visual_result}(a) is predicted incorrectly. 
By coupling with the CAM path, as shown in Figure~\ref{fig:visual_result}(d), the offsets of the crosswalk can be accurately estimated. 
In Figure~\ref{fig:visual_result}(e), the crosswalk region has a low (dark) motion confidence using only the OBJ path, while the confidence becomes much higher (brighter) in Figure~\ref{fig:visual_result}(f) with both paths.

\section{Conclusion} \label{sec:5}
We presented an efficient yet effective frame alignment method for the online-VSR problem.
Unlike existing online-VSR methods which compute the motion fields of each frame separately with a weight-sharing offset estimation module, we developed a novel temporal motion propagation (TMP) module, which propagates the motion field of the previous frame to the current frame, reducing largely the computational cost of offset estimation.
We further presented a motion confidence weighted fusion (MCRF) strategy to more effectively exploit the well-matched and complementary features for video enhancement.
The proposed method achieved not only higher PSNR/SSIM measures but also faster inference speed than the current state-of-the-art online-VSR methods. It can also be extended to other non-online VSR tasks. 


\ifCLASSOPTIONcaptionsoff
  \newpage
\fi

\bibliographystyle{IEEEtran}
\bibliography{IEEEabrv,refer}

\end{document}